# Normal Learning in Videos with Attention Prototype Network

Chao Hu, Fan Wu, Weijie Wu, Weibin Qiu, Shengxin Lai

Unicom (Shanghai) Industry Internet Co., Ltd.

{name}@chinaunicom.cn

## Abstract

*Frame reconstruction (current or future frame) based on Auto-Encoder (AE) is a popular method for video anomaly detection. With models trained on the normal data, the reconstruction errors of anomalous scenes are usually much larger than those of normal ones. Previous methods introduced the memory bank into AE, for encoding diverse normal patterns across the training videos. However, they are memory consuming and cannot cope with unseen new scenarios in the testing data. In this work, we propose a self-attention prototype unit (APU) to encode the normal latent space as prototypes in real time, free from extra memory cost. In addition, we introduce circulative attention mechanism to our backbone to form a novel feature extracting learner, namely Circulative Attention Unit (CAU). It enables the fast adaption capability on new scenes by only consuming a few iterations of update. Extensive experiments are conducted on various benchmarks. The superior performance over the state-of-the-art demonstrates the effectiveness of our method.* **Our code is avail-able at** *https://github.com/huchao-AI/APN/.*

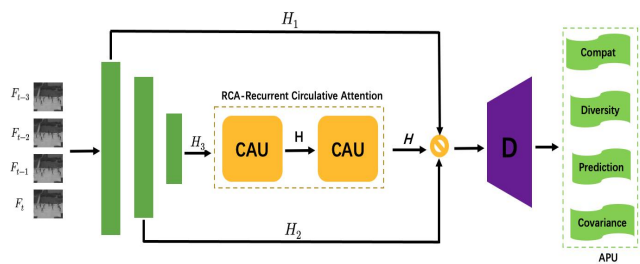

Figure 1: An overview of our approach. (1) We design an Attention Prototype Unit (APU) to learn a pool of prototypes for encoding normal feature; (2) Circulative Attention methodology is introduced to formulate the APU as normal latent space learner – Cirulative Attention Unit (CAU). It improves the feature learning capacity by extracting contextual information in more effective and efficient way. And RCA is equipped with the two CAU to enable network robust and effictive. Better viewed in color.

## 1. Introduction

Video anomaly detection (VAD) refers to the identification of behaviors or appearance patterns that do not conform to the expectation [2, 3, 5, 28]. Recently, there is a growing interest in this research topic because its key role in surveillance for public safety, *e.g.* the task of monitoring video in airports, at border crossings, or at government facilities becomes increasingly critical. However, the 'anomaly' is conceptually unbounded and often ambiguous, making it infeasible to gather data of all kinds of possible anomalies. Anomaly detection is thus typically formulated as an unsupervised learning problem, aiming at learning a model to exploit the regular patterns only with the normal data. During inference, patterns that do not agree with the encoded regular ones are considered as anomalies.

Deep Auto-Encoder (AE) [38] is a popular approach for video anomaly detection. Researchers usually adopt AEs to model the normal patterns with historical frames and to reconstruct the current frame [11, 31, 39, 4, 40, 1] or predict the upcoming frame [22, 34, 24, 26, 10]. For simplicity, we refer to the two cases as frame prediction. Since the models are trained with only normal data, higher prediction errors are expected for abnormal (unseen patterns) inputs than those of the normal counterparts. Previously, many methods are based on this assumption for anomaly detection. However, this assumption does not always hold true.

On the one hand, the existing methods rely on large volumes of normal training data to model the shared normal patterns. These models are prone to face the 'overgeneralizing' dilemma, where all video frames can be predicted well, no matter they are normal or abnormal, owing to the powerful representation capacity of convolutional neural networks (CNNs) [37, 10]. Previous approaches [37, 10] proposed to explicitly model the shared normal patterns across normal training videos with a memory bank, for the propose of boosting the prediction of normal regions in frames while suppressing the abnormal ones. However, it is extremely memory-consuming for storing the normal patterns as memory items across the whole training set.

*Corresponding authors

To tackle this limitation, we propose to encode the normal latent space in an attention manner, which is proven to be effective in representation learning and enhancement [46, 20, 13]. A normalcy learner, named as Attention Prototype Unit (APU), is developed to be easily incorporated into the AE backbone. It takes the encoding of consecutive normal frames as input, then learns to mine diverse normal flexible as compact prototypes. More specifically, we apply a novel attention operation on the AE encoding map, which assigns a normalcy weight to each pixel location to form a normalcy map. Then, prototypes are obtained as an ensemble of the local encoding vectors under the guidance of normalcy weights. Multiple parallel attention operations are applied to generate a pool of prototypes. With the proposed compactness and diverseness feature reconstruction loss function, the prototype items are trained to represent diverse and compact complexity of the shared normal patterns in an end-to-end fashion. Finally, the AE encoding map is aggregated with the normalcy encoding reconstruct-ed by prototypes for latter frame prediction.

On the other hand, the normal patterns appearing in various scenes differ from each other. For instance, a person running in a walking zone is regarded as an anomaly, while this activity is normal in the playground. Previous methods [22, 10] assume the normal patterns in training videos are consistent with those of test scenes in the unsupervised setting of VAD. However, this assumption is unreliable, especially in real world applications where surveillance cameras are installed in various places with significantly different scenarios. Therefore, there is a pressing need to develop an anomaly detector with adaption capability. To this end, [37] defines a rule for updating items in the memory bank based on a threshold to record normal patterns and ignore abnormal ones. However, it is impossible to find a uniform and optimal threshold for distinguishing the normal and abnormal frames under various scenarios.

In this work, we approach this problem from a new perspective, motivated by [25], consecutive sparse attention to replace the single layer dense attention in the non-local network. Without loss of generality, we use two consecuive recurrent attention modules, in which one only has sparse connections ($H+W$-1) for each position in the feature maps. The circulative attention module aggregates contextual information in horizontal and vertical directions. By serially stacking two recurrent attention modules, it can collect contextual information from all pixels. The above decomposition strategy greatly reduces the complexity in time and space from $O((H \times W) \times (H \times W))$ to $O((W \times H) \times (W \times H))$.

Circulative Attention Unit (CAU), for the goal of learning to learn the remarkable normalcy features. So the latent space can be very robust when in the complex scene. At present, the whole network [25] may lead to the 'over generalizing' problem, because the latent feature can not respresent whole normalcy scene. CAU pass through one recurrent attention module to collect the contextual information in horiontal and vertical directions. Then , by feading the produced feature maps from the first attention module to the other one. An overview of our approach is presented in Fig. 1.

We summarize our contributions as follows: i) We develop an Attention Prototype Unit (APU) for learning to represent diverse and various patterns of the normal data as prototypes. An attention operation is thus designed for aggregating the normal various to form prototype items. The whole process is differentiable and trained end-to-end.
ii) We introduce circulative attention into our backbone and improve it as a robust normalcy learner – Circulative Attention Unit (CAU). It effectively endows the model with the fast robust capability by consuming only a few parameters and update iterations. iii) Our APUbased AE achieves new state-of-the-art (SOTA) performance on various unsupervised anomaly detection benchmarks. In addition, experimental results validate the robust capability of our CAU in the complex scene.

## 2. Related Work

**Anomaly Detection.** Due to the absence of anomaly data and expensive costs of annotations, video anomaly detection has been formulated into several types of learning problems. For example, the unsupervised setting assumes only normal training data [19, 27, 23], and weakly-supervised setting can access videos with video-level labels [43, 53, 28]. In this work, we focus on the unsupervised setting, which is more practical in real applications. For example, the normal video data of surveillance cameras are easily accessible for learning models describing the normality. Earlier methods, based on sparse coding [7, 51, 23], markov random field [14], a mixture of dynamic textures [30], a mixture of probabilistic PCA models [15], *etc.*, tackle the task as a novelty detection problem [28]. Latter, deep learning (CNNs in particular) has triumphed over many computer vision tasks including video anomaly detection (VAD). In [27], Luo *et al.* propose a temporally coherent sparse coding-based method which can be mapped to a stacked RNN framework.

Recently, many methods leverage deep Auto-Encoder (AE) to model regular patterns and reconstruct video frames [11, 31, 39, 4, 40, 1]. Multiple variants of AE have been developed to cooperate spatial and temporal information for video anomaly detection. In [26, 6], the authors investigate Recurrent Neural Network (RNN) and Long Short Term Memory (LSTM) for modeling regular patterns in se-quential data. Liu *et al*. [22] propose to predict the future frame with AE and Generative Adversarial Network (GAN). They assume anomalous frames are unpredictable in the video sequence. It has achieved superior performance over previous reconstruction-based methods. However, this kind of methodology suffers from the 'over-generalizing' problem that sometimes anomalous frames can also be predicted well (*i.e.* small prediction error) as normal ones.

Gong *et al*. (MemAE) [10] and Park *et al*. (LMN) [37] introduce a memory bank into the AE for anomaly detection. They record normal patterns across training videos as memory items in a bank, which brings extra memory cost. While we propose to learn the normalcy with an attention mechanism to measure the normal extent. The learning procedure is fully differentiable and the prototypes are dynamically learned with the benefits of adapting to the current scene spatially and temporally, compared with querying and updating the memory bank with pre-defined rules for recording rough patterns cross the training data in [10, 37]. Moreover, the prototypes are automatically derived based on the real-time video data during inference, without referencing to the memory items collected from the training phase [10, 37]. For adaption to test scenes, Park *et al*. [37] further expand the update rules of the memory bank by using a threshold to distinguish abnormal frames and record normal patterns. However, it is impossible to find a uniform and optimal threshold for distinguishing the normal and abnormal frames under various scenarios. On the contrary, we introduce the cov-restrain technology into our APU module to enable the fast adaption capacity to a new scenery.

**Attention Mechanisms.** Attention mechanism [48, 49, 13, 42, 16, 50, 9, 52, 20] is widely adopted in many computer vision tasks. Current methods can be roughly divided into two categories, which are the channel-wise attention [49, 13, 42, 50] and spatial-wise attention [42, 52, 49, 16, 9]. SENet [13] designs an effective and lightweight gating mechanism to self-recalibrate the feature map via channel-wise importance. Wang *et al*. [48] propose a trunk-and-mask attention between intermediate stages of a CNN. However, most prior attention modules focus on optimizing the backbone for feature learning and enhancement. We propose to leverage the attention mechanism to measure the normalcy of spatial local encoding vectors, and use them to generate prototype items which encode the normal patterns.

**Context information aggregation.** In addition, some works aggregate the contextual information to augment the feature representation dilation convolutions to capture contextual information. Dense ASPP brought dense conections into ASPP to generate features with various scale. DP-C utilized architecture search techniques to build multi-scale architectures for semantic segmentation. PSPNet utilized pyramid pooling to aggregate contextual information. GCN utilized global convolutional module utilized global pooling to harvest context information for repre sentations. Recently, Zhao *et al*. proposed the point-wise spatial attention network which uses predicted attention map to guide contextual information collection. Liu *et al*. utilized RNNs to capture long-range dependecies. Conitional random field (CRF) , Markov random field (MRF) are also utilized to capture longrange dependencies for semantic segmentation. However, in this work, we utilized skip connection in AE, so it can robust the capacity of extracting features, CAU is made with two attention modules to aggregate the context information.

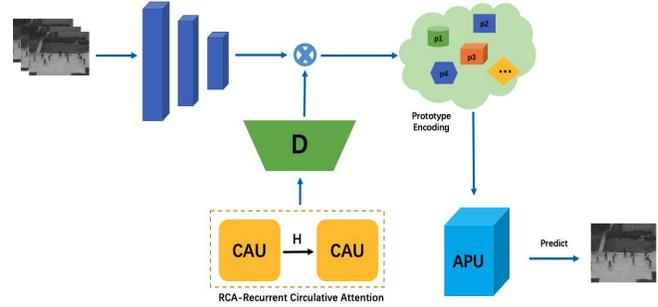

Figure 2: The framework of APU-based model. The proposed Attention Prototype Unit (APU) is plugged into an Auto Encoder (AE) to learn prototypes for encoding normal features. The prototypes are obtained from the AE encoding with the guidance of normalcy weights and the normalcy weights of the AE encoding are generated in a fully differentiable attention manner. Then an normalcy encoding map (blue color) is reconstructed as an encoding of learned prototypes. It is further aggregated with the AE encoding map for latter frame prediction.

## 3. Method

In this section, we elaborate the proposed method for VAD. First, we describe the learning process of normal learning in the Attention Prototype Unit (APU) in Sec. 3.1, and we explain the objective functions of the framework in Sec. 3.2. Then in Sec. 3.3, we present the details of the latent space normalcy learner. Finally in Sec. 3.4, we detail the training and testing procedures of our VAD framework.

### 3.1. Attention Prototype Unit

The framework of the APU-based AE is shown in Fig. 2. APU is trained to learn and compress normal features of real time sequential information as multiple prototypes and enrich the input AE encoding with normal flexible information. Note that, APU can be plugged into different places (with different resolutions) of the AE. We conduct ablation studies in Sec. 4.4 to analyze the impact of APU position. Let's first consider an AE model takes as inputs the To bserved video frames ($I_k-T+1$, $I_k-T+2$, ..., $I_k$), simplified as *v*, Then the selected hidden encoding of AE is feed ward into our APU, Finally, the output encoding of is run through the remaining AE layers (after APU) for predicting upcoming ground-truth frame $Y_k = Y_k+1$. We denote the frame sequence as an input&output pair ( $X_k$, $X_k$) of the k-th moment.

The forward pass of APU is realized by generating a pool of normal prototypes in a fully differentiable attention manner, then reconstructing a normalcy encoding by retrieving the prototypes, and eventually aggregating the input encoding with the normalcy encoding as the output. The whole process can be broken down into 3 sub-processes, which are *Attention*, *Distingusih*, *Ensemble* and *Retrieving*.

Currently, the t-th input encoding map from AE is first extracted, viewed as N = w * h vectors of c dimensional, $x_t^1, x_t^2, ..., x_t^N$. In the sub-process of Attention, a quantity of M attention mapping functions $\{\psi_m : R_c \rightarrow R_l\}_{m=1}^M$ are employed to assign normalcy weights to encoding vectors, On each pixel location . On, the normalcy weight measures the n-ormalcy extent of the encoding vector. Here, denotes the m-th normalcy map, generated from the m-th attention function. Then one unique prototype is derived as an ensemble of N encoding vectors with normalized normalcy weights in sub-process Ensemble as:

$$p_t^m = \sum_{n=1}^{N} \frac{w_t^{n,m}}{\sum_{n'=1}^{N} w_t^{n',m}} x_t^n. \qquad (1)$$

Similiary, M prototypes are derived from multiple attention functions to from a prototype pool, In the Retrieving subprocess, input encoding vectors from the AE encoding map are used as queries to retrieve relevant items int the prototype pool for recontructing a normalcy encoding. For every obtained normalcy vector, this proceeds as :

$$\tilde{x}_t^n = \sum_{m=1}^{M} \beta_t^{n,m} p_t^m, \qquad (2)$$

where P denotes the relevant score between the n-th encoding vector X and the m-th prototype item. The obtained normalcy map is aggregated with the original encoding as the final output using channel-wise sum operation. The key idea is to enrich the AE encoding with the normalcy information to boost the prediction of normal parts of video frames while suppressing the abnormal parts.

Finally, in the Distingusih sub-procees module, the input features come from the AE modules. The goal is to distingu-sh different distribution of the AE encoding vectors. So that introducing Covariance Matrix, it can be taken of various distribution, and the input encoding vectors will be independence. So this operation can be confirmed that all the prototypes are not matter with each other. For very obtained normalcy encoding vector, this proceed as:

$$C = (c_{ij})_{n \times n} = \begin{pmatrix} c_{11} & c_{12} & \cdots & c_{1n} \\ c_{21} & c_{22} & \cdots & c_{2n} \\ \cdot & \cdot & & \cdot \\ \cdot & \cdot & & \cdot \\ \cdot & \cdot & & \cdot \\ c_{n1} & c_{n2} & \cdots & c_{nn} \end{pmatrix} \qquad (3)$$

The output will be imbedding to predict the feature video frame.

### 3.2. VAD Objective Functions

In this section, we present the objective functions in the pipeline, which enable the prototype learning for normalcy representation, feature reconstruction for normalcy enhanced encoding, and frame prediction for anomaly detection. To train our model, the overall loss function L consists of a feature reconstruction term and a frame prediction term. These two terms are balanced by weight as:

$$\mathcal{L} = \mathcal{L}_{\text{fra}} + \lambda_1 \mathcal{L}_{\text{fea}}. \qquad (4)$$

**Frame Prediction Loss** is formulated as the L2 distance between ground-truth and prediction:

$$\mathcal{L}_{\text{fra}} = \|\hat{y}_t - y_t\|_2. \qquad (5)$$

**Feature Recontruction Loss** is designed to make the learned normal prototypes have the properties of compactness and diversity. It has three terms, aiming at the two properties respectively, and is written as:

$$L_{fea} = L_c + \lambda_2 L_d + \lambda_3 L_{cov} \qquad (6)$$

where the parameters is the balanced weights. The compactness term is for construction of normalcy encoding with compact prototypes. It measures the mean L2 distance of input encoding vectors. And diversity term means the relevant score of the most relevant vectors. With the argmax, can be used to promote the diversity among prototypes items by pushing the learned prototypes away from each other. And we used covariance to confirm the independence of the encoding vectors. And also it can be push all learned prototype far away, so that the learned normalcy prototypes is max distance and independent. All trained tricks math are written as:

$$L_c = \frac{1}{N}\sum_{n=1}^{N}|x_t - p_t| \qquad (7)$$

$$L_d = \frac{2}{M(M-1)}\sum_{m}^{M}\sum_{m^*}^{M}\|-(p_m - p_{m^*})\| \qquad (8)$$

$$L_{cov} = \frac{2}{M(M-1)}\sum_{m=1}^{M}\sum_{m^*}^{M}cov\|p_m - p_{m^*}\| \qquad (9)$$

Taking benefits of above three terms, the prototypes latent space are encouraged to encode compact and diverse normalcy latent features for normal frame prediction.

### 3.3. Circulative Attentions

To module full image dependencies over local feature representations using lightweight computations and memory, we introduce a recurrent attention module. The recurrent attention module collect contextual information in all axis directions to enhance pixel-wise representative capability.

We first introduced two convolution layers 1×1 filters on h to generate two feature maps Q and K, and channel C is the number of all channel, which is for dimension reduction. After obtaining feature maps Q and K, we further generate attention maps A via affinity operation. At each position $u$ in the spatial dimension of feature maps Q, we can obtain a vector Q. Mean while set Q e-tracting features vectors from K whick are in the same row or column with position $u$. And position u is below to index of image level.

Given the *X*, we first apply a convolutional layer to obtain the feature maps H of dimension reduction, then, the feature maps H are fed into the Circulative Attention module to generate new feature maps **H** which aggregate contextual information together for each pixel in its recurrent attention path. The **H** only aggregate the contextual information in all directions which are not powerful for other tasks. To obtain richer and denser context information, we feed the feature maps **H** into the circulative attention module again and output feature maps **H**. Thus, each position in feature maps **H** actually gathers the information from all pixels. Two circulative attention module before and after share the same parameters to avoid adding too many extra parameters. So we name this recurrent structure as circulative attention unit(CAU) module.

Then, we concatenate the dense contextual feature **H** with the local representation feature **X**. It is follwed by one or several convolution for feature fusion. Finally, the fused features are fed into the special task layers to predict the final result.

**Recurrent Circulative Attention**. Despite a circulative attention can capture contextual information in all directions, the connections between one pixel and its around ones that are not in the inner circulative path are still absent. To tackle this problem, we innovatively introduce a RCA operation based on the circulative attention. The RCA module can be unrolled into R loops. In the first loop, the CAU takes the feature maps H extracted from a CNN model as input and output the feature maps **H**, where **H** and H are with the same shape. As shown Fig. 3, the RCA module is equipped with two loops(*R*=2) which is able to harvest full-image contextual information from all pixels to generate new feature mpas with dense and rich contextual information.

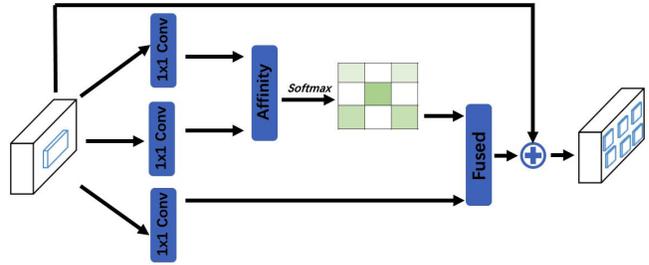

Figure 3. The details of CAU module

We denote A and A as the attention maps in loop 1 and loop 2, respectively. Since we interested only in contextual information spreads in spatial dimension rather than in channels, the convolutional layer with 1×1 filters can be view as the identical connection. In addition, the mapping function from position $x$, $y$ weight **A** which stands for the index of $i$, $x$, $y$ is defined as $\mathbf{A} = f(A, x, y, x, y)$. For any position $u$ at the feature maps **H** and any position $\theta$ at the feature maps **H** there is actually a connection in case of $R = 2$. For the case $\theta$ and $u$ are in the same directions.

In general, our RCA module makes up of the deficiency of recurrent circulative attention that obtain the dense contextual information from all pixels. Compared with circulative attention, the RCA moudle ($R = 2$) does not bring extra parameters and can achieve better performance with the cost of minor computation increment.

### 3.4. Video Anomaly Detection Pipeline

We first explain the details of the whole network architecture and how anomaly scores are generated. Then we describe the training and testing phases of our framework.

**Network Architecture Details.** Our framework is implemented as a single end-to-end network illustrated in Fig. 2. We adopt the same network architecture in [22, 37] as the backbone of AE to facilitate a fair comparison. In the CAU module, M attention mapping functions are implemented as fully connected layers to generate a series of normalcy maps and further to form a pool of flexible prototypes. The output encoding of APU is put forward through the decoder of AE for frame prediction. In addition, the APU module is consisted of fore restraint, such as compact, diversity, reconstruction, covariance. The details are explained below.

**Anomaly Score.** To better quantify the anomalous extent of a video frame during inference, we investigate the two cues of feature reconstruction and frame prediction. Since the normal flexible items in the normal prototype pool are learned to encode the compact representations of the normal encoding as in Eq. 7, during inference, an anomaly score can be naturally obtained by measuring the compactness error of feature reconstruction term. And the two variable denote the input encoding map and the flexible prototype pool of the t-th moment, respectively. And we introduce covariance of the prototype pool to confirm the learned prototype independence. As in previous methods [22, 10,

37], frame prediction error is also leveraged as an anomaly descriptor. Thus we obtain above two kinds of anomaly scores and combine them with a balance weight. Besides our backbone is besed on RestNet-50 and remove the last two down-sampling operations.

**Training Phase.** Adamw with mini-batch is used for training. For anomalous event detection, the channels of input datas are catennated to network feed-forward, so that this way can be instead of the flow methods. The initial model can be first trained in the training dataset, and fixed the backbone to optimize the APU and prediction loss and reconstruction loss. By these method, the initial parameters is obtained, so the model with several epochs from videos of diverse scenes can be adapted for any scene.

**Testing Phase.** In the testing phase, given a new test sequence, we simply use the first several frames of the sequence to construct input&output frame pairs for updating model parameters. The same procedure is used in the training phase. The updated model is used for detecting anomalies afterwards.

## 4. Experiments

### 4.1. Problem Settings, Datasets and Setups

**Problem Settings.** For better evaluating the effectiveness of our approach, we follow two anomaly detection problem settings, which are the unsupervised setting and different training sets. The first one is widely adopted in existing literature [37, 10, 22, 19, 23, 27], where only normal videos are available during training. The trained models are used to detect anomalies in test videos. Note that the scenarios of test videos are seen during training in this setting. The second one, for evaluating the model robust, is based on collecting training and testing videos from different datasets to make sure the diversity of scenarios during training and testing. This setting is also called 'cross-dataset' testing in [25]. In summary, the first setting challenges the approaches for how well they can perform under one fixed camera, while the latter setting examines the adaption capability, when given a new camera. We believe above settings are essential for evaluating a robust and practical anomaly detection method.

**Datasets.** Four popular anomaly detection datasets are selected to evaluate our approach under different problem settings. 1) The UCSD Ped1 & Ped2 dataset [19] contains 34 and 16 training videos, 36 and 12 test videos, respectively, with 12 irregular events, including riding a bike and driving a vehicle. 2) The CUHK Avenue dataset [23] consists of 16 training and 21 test videos with 47 abnormal events such as running and throwing stuff. 3) The Shanghai Tech dataset [27] contains 330 training and 107 test videos of 13 scenes. 4) The UCF Crime dataset [43] contains normal and crime videos collected from a lot of real world surveillance cameras where each videos come from video comes from a different scene. We use the 950 normal videos from this dataset for evaluating the robust of training, then test the model on other datasets in the cross-dataset testing as in [25].

**Evaluation Metrics.** Following prior works [22, 26, 30], we evaluate the performance using the area under ROC curve (AUC). ROC curve is obtained by varying the threshold for the anomaly score for each frame-wise prediction.

**Implementation Details.** Input frames are resized to the reesolution of 256 x 256 and normalized to the range of [1, 1]. During the AE pre-training, the model is trained with the learning rate as 0.0001 and batch size as 4. In the default setting, CAU is plugged into the AE after the third CNN layer counting backwards, with the encoding feature map of resolution 256 x 256 x 128. Training epochs are set to every 30 epochs to saved on Ped1, Ped2, Avenue and Shanghai Tech, respectively, and the total epochs is 1000. During the process of training phase, the AE backbone can be frozen, only the target model APU is trained. The learning rate of the update iteration of the APU parameter set $\theta$ is set to 0.00001 for 1000 training epochs. The mini-batch is set as 10 episodes, and the learning rate of step size $\alpha$ is 0.00001. The balance weights in the objective functions are set as $\lambda 1 = 1$, $\lambda 2 = 0.01$, $\lambda 3 = 0.01$. The desire margin $\gamma$ in feature diversity term is set to 1. Finally, the hyper parameter $\lambda s$ is set to 1. The experiments are conducted with four Nvidia RTX-3090 GPUs.

### 4.2. Comparisons with SOTA Methods

**Evaluation under the unsupervised setting**. We first perform an experiment to show that our proposed backbone architecture is comparable to the state-of-the-arts. Note that this sanity check uses the standard training test setup (training set and testing set are provided by the original datasets), and our model can be directly compared with other existing methods. Table 1 shows the comparisons among our proposed architecture and other methods when using the standard unsupervised anomaly detection setup on several anomaly detection datasets. MemAE [10] and LMN [37] are most-related methods to our approach. They learn a large memory bank for storing normal patterns across the training videos. While we propose to learn a few flexible normal prototypes conditioned on input data, which is more memory-efficient. The superior performance also demonstrates the effectiveness of our APU module. On ped1 and Shanghai Tech, AUCs of our approach are lower than those of rGAN [25]. This is reasonable because the model architecture of rGAN is more complicated. rGAN uses a ConvLSTM to retain historical information by stacking AE several times. However, we only apply a single AE. Additionally, we trained the best learned model with ResNet model using both training and validation datasets.

Table 1: Quantitative comparison with state-of-the-art methods for anomaly detection. We measure the average AUC (%) on UCSD Ped1&Ped2 [19], CUHK Avenue [23], and ShanghaiTech [27] in the unsupervised setting. Numbers in bold indicate the best performance and underscored ones are the second best.

| Methods | Ped1 | Ped2 | Avenue | Shanghai |
|---|---|---|---|---|
| MPPCA [14] | 59.0 | 69.3 | - | - |
| MPPC+SFA [14] | 68.8 | 61.3 | - | - |
| MDT [30] | 81.8 | 82.9 | - | - |
| MT-FRCN [12] | - | 92.2 | - | - |
| Unmasking [45] | 68.4 | 82.2 | 80.6 | - |
| SDOR [35] | 71.7 | 83.2 | - | - |
| ConvAE [11] | 75.0 | 85.0 | 80.0 | 60.9 |
| TSC [27] | - | 91.0 | 80.6 | 67.9 |
| StackRNN [27] | - | 92.2 | 81.7 | 68.0 |
| Frame-Pred [22] | 83.1 | 95.4 | 85.1 | 72.8 |
| AMC [34] | - | 96.2 | 86.9 | - |
| rGAN* [25] | 83.7 | 95.9 | 85.3 | 73.7 |
| rGAN [25] | 86.3 | 96.2 | 85.8 | **77.9** |
| MemAE [10] | - | 94.1 | 83.3 | 71.2 |
| LMN [37] | - | 97.0 | 88.5 | 70.5 |
| MPM[54] | 85.1 | 96.9 | 89.5 | 73.8 |
| Ours w/o APU. | 83.2 | 95.1 | 84.0 | 66.7 |
| Ours w APU. | **87.2** | **98.2** | **88.7** | 77.5 |

Table 2. Comparison with state-of-the-arts on Ped2.

| Method | Backbone | multi-scale | Ped2 |
|---|---|---|---|
| ConvAE [11] | ResNet-100 | No | 85.3 |
| AMC[34] | Xception-50 | Yes | 96.5 |
| MPN[54] | Xception-50 | Yes | 96.9 |
| Ours | ResNet-50 | No | 96.5 |
| Ours | ResNet-50 | Yes | 98.2 |

**Evaluation under the robust setting.** To demonstrate the scene adaption capacity of our approach, we conduct cross-dataset testing by multi-scale training on the training set of Ped2 and normal videos of Ped1, The comparison results are reported in Table 2. As we can see, Results of other state-of-the-art competitive results on Ped2 validation set are summarized in Tab. 2. We provide these results for reference and emphasize that these results should not be simply compared with our method, since these methods are trained on diferent (even larger) training sets or different basic network. Among these approaches, MPN[11] and AMC[34] adopt the same backbone and multi-scale testing strategy, ConvAE [11] use a more stronger backbone, for pre-training beyond the training set of Ped2. The results show that the proposed APN with multi-scale testing still outperforms all these strong baselines. Feature reconstruction with the multi-scale of CAU moudle based on prototypes largely boosts the robustness of anomaly detection with frame prediction. Furthermore, 4 5% gain can be achieved with CAU. The performance of our APU-based AE is superior and comparable to the SOTA learner methods, with a significantly faster adaption and inference speed. We prov-ide more tailed model complexity and inference speed in Sec. 4.3.

Table 3: Analysis on the model complexity and inference speed of various SOTA methods. The inference speed information is collected by running the official implements on a single Nvidia RTX 3090 GPU on a machine with 8 CPU cores of E5-2650 v4-@2.20GHz and 24 G memory.

| Methods | Parameters (M) | FPS |
|---|---|---|
| rGAN [25] | 19.0 | 2.1 |
| MemAE [10] | 6.2 | 86.7 |
| LMN [37] | 15.0 | 126.3 |
| Ours | 14.68 | 122.3 |

### 4.3. Model Complexity and Inference Speed

With a single Nvidia RTX-2080Ti GPU, our model can run at 166.8 FPS. Note that our APU module only consumes 14.68K extra parameters (with 10 prototypes). Although the parameter size of MemAE [10] is smaller than that of ours, the large memory bank used in MemAE leads to a time consuming read operation, so as the whole inference procedure. Apart from model parameters, our model does not need extra memory space for prototypes, which can be viewed as latent feature vectors. Moreover, the inference speed of the method is almost 60 × faster than rGAN[25]. The update iteration for scene rubst of RCA model ($R = 2$) takes only 0.008 seconds (122.3 FPS). This is almost so fast than most methods. The fast inference speed makes our model more favorable in real world applications. The comparison result are reported in Table 3.

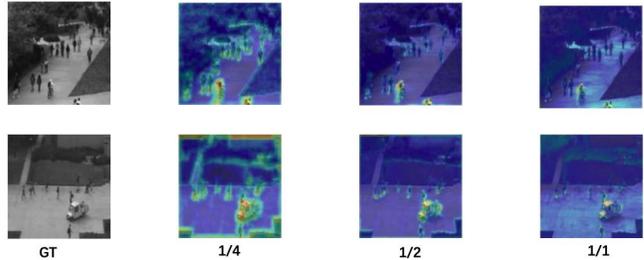

Fig 3: Visualization of AE encoding activation maps from the perspective o L2 norm. GT stands for ground truth frame and other columns denotes the different corresponding attention maps of input images resolution.

### 4.4. Ablation Studies

**Model Component Analysis.** We first analyze the effectiveness of APU. We set $M = 10$ as the default number of the attention mapping functions in APU. The results are listed in Table 4. It is clear that the overall performances on various benchmarks are boosted with our APU by a large margin. We also visualize some example prediction error maps as well as the normalcy maps in APU in Fig. 5. To better analyze the learned normalcy maps, we aggregate all $M$ maps by the sum operation as in Fig. 6 (c). The normalcy maps encode diverse normal attributes of the scenes such as roads, grasses, and buildings, shown in the columns of (d) ~ (f). Furthermore, the weights in suspicious regions are far smaller than those in other parts of the map, indicating that the normal patterns are well encoded as prototypes.

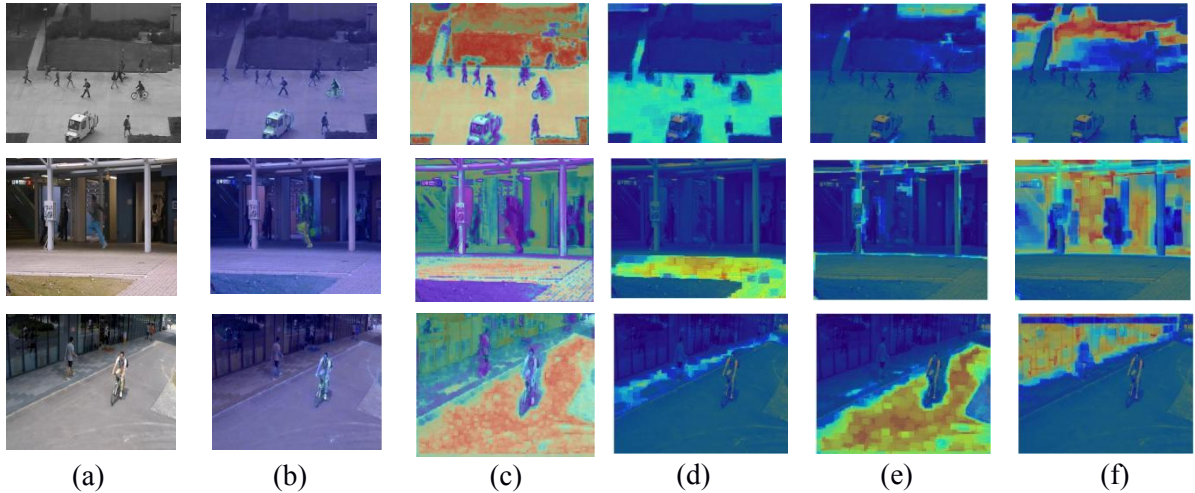

(a) (b) (c) (d) (e) (f)

Figure 5: Visualization of some examples of test cases and APU normalcy maps. The groups of pictures in different columns denote (a) ground-truth frame, (b) error map, (c) sum of normalcy map in APU, (d) ~ (f) various normalcy map, respectively.

Table 4: AUC analysis of the designed APU module. In the table, FR and FP stand for anomaly scores derived from the Feature Reconstruction and the Frame Prediction, respectively.

| Setting | Shanghai | Avenue | Ped2 | Ped1 |
|---|---|---|---|---|
| AE baseline (FP) | 67.8 | 84.9 | 95.6 | 84.3 |
| AE with APU (FP) | 75.1 | 86.2 | 93.6 | 85.5 |
| AE with APU (FR) | 75.9 | 87.5 | 95.6 | 75.1 |
| AE with APU (FP & FR) | **77.5** | **88.7** | **98.2** | **87.2** |

Table 6: Analysis on the plugging spot of the APU module. The resolution is divided by the resolution of input images (256×256).

| Resolution | **1/1** | 1/2 | 1/4 |
|---|---|---|---|
| AUC | **90.2** | 87.8 | 85.7 |

Table 7: AUC analysis on the quantity of prototypes in APU.

| Number | 1 | 5 | **10** | 20 | 40 |
|---|---|---|---|---|---|
| FR | 88.59 | 91.59 | **94.69** | 89.28 | 85.47 |
| FP | 95.66 | 96.45 | **95.23** | 95.90 | 95.61 |
| Overall | 95.24 | 95.57 | **98.20** | 96.43 | 95.74 |

**APU Resolution Analysis.** To investigate the effect of the plugging spot of the APU module, we carry out experiments on four positions with encoding maps of different resolutions. The results are listed in Table 6. The AUC results Are derived from the feature reconstruction anomaly score on Ped2 dataset. The performance increases along with the resolution. We visualize the activation map of the encoding using the L2-norm of the spatial encoding vectors in Fig. 3. The higher the activation value, the more information is included in the encoding vector. We find that in the higher resolution layers of AE, more anomaly cues are included, which is beneficial for measuring the anomalous extent with the feature reconstruction.

**Prototype Quantity Analysis.** To encode the normal dynamics as prototypes, we propose to leverage multiple att-ention mapping functions for measuring the normalcy of encoding vectors and deriving prototypes as ensembles of the vectors. The number of the attention functions, also denoting the quantity of prototypes, serves as the up-bound of the diverse prototypes needed in one scenario. Experimental results on Ped2 are in Table 7. Based on the results, $M = 10$ is an appropriate number of required prototypes. With the number increasing, more noise information is involved and the diversity of prototype items can not be guaranteed, leading to a drastic decline of the performance.

## 5. Conclusion

In this work, we have introduced a prototype learning module to explicitly model the normal flexible in video sequences with an attention mechanism for unsupervised anomaly detection. The prototype module is fully differentiable and trained in an end-to-end manner. Without extra memory consumption, our approach achieves SOTA performance on various anomaly detection benchmarks in the unsupervised setting. In addition, we improve the prototype module as a rebustness normalcy learner with the robust learning technology. Extensive experimental evaluations demonstrate the efficiency of the scene adaption approach.

**Acknowledgments.** This work was supported by the National Natural Science Foundation of China (Grants Nos. 62072244, 61972204, 61906094), the Natural Science Foundation of Jiangsu Province (Grant No. BK20190019).